# Model Adaption Object Detection System for Robot

Jingwen Fu, Licheng Zong, Yinbing Li, Ke Li, Bingqian Yang, Xibei Liu

Xi'an Jiaotong University, Xi'an,Shaanxi, 710049, P. R. China
E-mail: fu1371252069@stu.xjtu.edu.cn

**Abstract:** Object detection for robot guidance is a crucial mission for autonomous robots, which has provoked extensive attention for researchers. However, the changing view of robot movement and limited available data hinder the research in this area. To address these matters, we proposed a new vision system for robots, the model adaptation object detection system. Instead of using a single one to solve problems, We made use of different object detection neural networks to guide the robot in accordance with various situations, with the help of a meta neural network to allocate the object detection neural networks. Furthermore, taking advantage of transfer learning technology and depthwise separable convolutions, our model is easy to train and can address small dataset problems.

**Key Words:** Deep Learning, Robot Vision System, Object Detection

## 1 Introduction

The object detection is a crucial capability for an autonomous robot in visual perception and interaction with the real world. Object detection is considered as a challenging task due to various factors. For example, it is always hard to collect enough data to train the robot vision system and the view of robot changes when it moves. Humans can easily find an object and locate it in our field of vision, but the current robot vision system, on the other hand, lags far behind the human performance level. One of the research hotspots today is the grasp detection. In this area, the researchers fix on using the object detection technology to help robots grasp objects on a table [1] [12] or assembly line [3], where the robots don't need to move and they just need to use their legs to grasp the object. However, in the real world, there are many robot tasks more difficult than it. For example, if we want to train a robot to pick up the rubbish and put it in the trash, the robot has to move close to the object before it grasps this object. Therefore, it is necessary for us to use object detection methods to guide the robots to get close to the object.

Nowadays, there are two main challenges in this task. First, the same as many other robots' tasks, it is really difficult for us to obtain enough data to train. As a result, the object detection method should be trained with small training data. What's more, the vision of the robots changes a lot during their moving process, which makes it difficult to choose a single object detection algorithm for every moment. To solve these problems, we proposed the Model Adaption Object Detection (MAOD) algorithm for this project.

Our method has two prominent advantages:
1) MAOD can adapt to different situations, so that it won't be impaired by the variational views of robot.
2) MAOD can achieve good performance after training with only a small training dataset, due to the transfer learning and depthwise separable convolutions technologies used here.

We will detailly illustrate these advantages and the reasons for us to achieve these advantages in secton 3.

## 2 Related Work

Object detection is a fundamental visual recognition problem in computer vision and has been widely studied in the past decades. Object detection techniques using deep learning have been actively studied in recent years. All the methods here can be divided into two categories:two-stage detectors and one-stage detectors. Two-stage detectors split the detection task into two stages:(i) proposal generation; and (ii) making predictions for these proposals, like R-CNN[8], SPP-Net [9], fast R-CNN [7] ,Faster R-CNN [19] and R-FCN[6], while one-stage detectors do not have a separate stage for proposal generation, like OverFeat [21],YOLO[16],SSD[14], YOLOv2[17],RetinaNet[13] and YOLOV3[18].

Even though the object detection algorithm can achieve good performance, employing these algorithms into a robot is still a question. The application of object detection in robot consists of two main aspects: moving object detection and grasp detection.

**moving object detection** Moving object detection aims to detect the object when the object is moving related to the camera. In 2009, Kundu et al.[11] successfully detect and track multiple moving objects like a person and other robots by detecting independently moving objects in image sequence from a monocular camera mounted on a robot . Later, a deep learning based block-wise scene analysis method equipped with a binary Spatio-temporal scene model[25] is used for moving object detection. At the same year, Chavez et al. propose a complete perception fusion architecture based on the Evidential framework to solve the Detection and Tracking of Moving Objects (DATMO) problem[4]. In 2019, Yahiaoui proposes a CNN architecture for moving object detection[22] using fisheye images that were captured in the autonomous driving environment.

**grasp detection** Grasp detection is used to provide information for the robot to grasp objections, which is a task robot usually come across in the reality. In 2017, Umar et al. presented an efficient framework, which used a novel architecture of hierarchical cascaded forests to perform recognition and grasp detection of objects from RGB-D images of real scenes[2]. However, in this work, Umar didn't use deep learning technology, a recent emerging, and powerful method for object detection. In recent years, many people have tried to apply this technology to grasp detection. Elio Ogas et al. used Convolutional Neural Networks for recognizing a selected production piece on a cluster[15], and it achieved a good performance using a common webcam as image input. In the same year, Lu et al. proposed a novel grasp detection model that was constructed to make a fairer evaluation on grasp candidate using the grasp path[5].

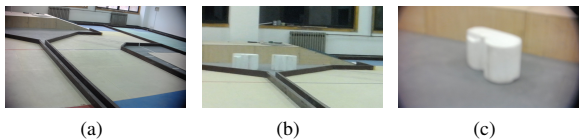

(a)          (b)          (c)

Fig. 1: Different situations the robot will come across from searching object to grasp the object. (a)Situation where there are no objects in the image. (b)Situation where there are remote objects in the image. (c)Situation where there is only one close object in the image

## 3 Model Adaption Object Detection System

### 3.1 Overall Structure

Fig 1 shows the different situations the robot will come across in this process. At first, the robot needs to search the object in a global view and here vision systems job is to determine whether objects exist in the view. For example, (a) is the situation, where there are no objects in the view and (b) shows the view, which contains two objects. After the robot finds the object, they need to locate this object roughly. Here, there are one or more objects in the view. Then, the robot will get closer to this object and the object becomes bigger and bigger in the robots view. At this time, only an object in the robots view and the robot need to get the fine location of the project so that it can obtain a good location to grasp it correctly. In order to address the variable views of the robot here, the MAOD will apply different operations according to the different views, so that we can choose better performance at a lower cost.

Under this condition, MAOD has the structure as shown in fig 2. First, there is an image acquisition subsystem to obtain the images from the camera on the robot and convert this image into a suitable form for our future analysis. Then, all the images are processed with the feature extractor, which is the convolution neural network with a fixed weight, and we can obtain the corresponding feature maps of these images. Afterward, the meta neural network can determine which situations the robot is in. If the robot is in the situation (a) in Fig 1, the system will give up the further processing of this image and turn to the image acquisition step to obtain a new image. If the robot is in the situation (b) in fig 1, the feature map of the images will be delivered to the rough object detection neural network and the neural network will extract the rough object information and send the information to the robot control system to influence the movement the action of the robot. And if the robot is in the situation (c) in fig 1, the robot will just apply the process the same as situation 2, except the fine object detection neural network used here, instead of the rough object detection neural network.

We have this flow path for the following reason:

1) if there are no objects, we won't use any object detection networks.
2) when the object is far away from us, we just need its center point to get closer to it, so we use the rough object detection neural network.
3) when the object is close enough to us, we can't ignore its shape, so at this point, we will use the fine object detection.

In the following, we will describe the image acquisition subsystem, meta neural network, and two object detection neural networks in detail.

### 3.2 image acquisition subsystem

The structure of image acquisition subsystem is shown in fig 3, from which we can obtain that when the robot moves to the specified position or triggers the specified condition and needs to detect the target area of the current imaging device, the motion control main chip STM32F407IG of the robot transmits the request for obtaining the object position to the PC through the RS-232C interface using the USART communication module. After the PC monitors and verifies the message received by the serial port module, it opens the image stream input of the corresponding image acquisition device, and stores the image stream input into the frame buffer. Having detected the picture appearing in the frame buffer, the image is used for object detection. Then the object coordinates are solved into the robot coordinate system in combination with the installation position calibration information of the imaging device, and the coordinate information is sent to the robot motion control main chip STM32F407IG through the RS-232C interface. After transmitting, the image stream input is closed and the frame buffer will be cleared.

### 3.3 Meta Neural network

The overall structure of the meta neural network is shown in fig 4, whose purpose is to determine which situations the robot is in. Therefore, it is just an image classifier here. The input of the network is the feature maps extracted by the feature extractor, which contains the low-level features of the image. Then multiple CNN layers are used to process the feature maps. All the weights in the CNN layers are pretrained using the Imagenet dataset. Given the 1024 layers feature obtained by CNN layers, the average global pool used here to convert these features into $1024 \times 1$ vector and linear layer farther convert this vector into $3 \times 1$ size. Each element in this vector corresponding to a situation of the robot. We choose the situation, whose corresponding value in the vector is the maximum, as the predicted situation for a robot. The loss function here is:

$$p_i = \frac{exp(o_i)}{\sum_j (exp(o_j))} \quad (1)$$

$$loss = -\sum_i (\alpha_i \cdot t_i \cdot log(p_i)) \quad (2)$$

where $o_i$ is the ith output of the network, $t_i$ denotes the ground true ,for example if the ground true is j, $t_j$ is equal to 1 and the others are 0, and the $\alpha_i$ is set manually to address the unbalance of the labels.

### 3.4 Rough Object Detection Neural Network

This network is designed to find the center of the object because when the object is far away from the robot, there is no need for the robot to get the exact information of the object. The robot just needs the information of the object's position to guide it to get close to this object. Fig 5 shows the overall structure here. First, the images are divided into n small parts. Each part is marked with a unique index (0,1,2,...,n) and at the end of the neural network, N scores are generated. Each score is corresponding to each part of the image with the same index. If the center of the object lays in part i of the images then score i is 1, otherwise, it is 0. In this way, we can roughly find the object in the images. Then, following the feature extractor layer, there is a dropout layer, which can make the element in the feature map become zero with a certain probability leading to the input more variable for the following layers. After the first CNN layers, the neural network forks into two ways. The global pool and linear layer are designed to find the meticulous information and the CNN layers will keep the space information. Then they emerge by adding the two features generated by each part and a following linear layer. The loss function here is the same as the meta neural network, except that there are more output here. The values of $alpha_i$ are more important here in the training process, because the probability of the object falling some parts of the image is higher than others.

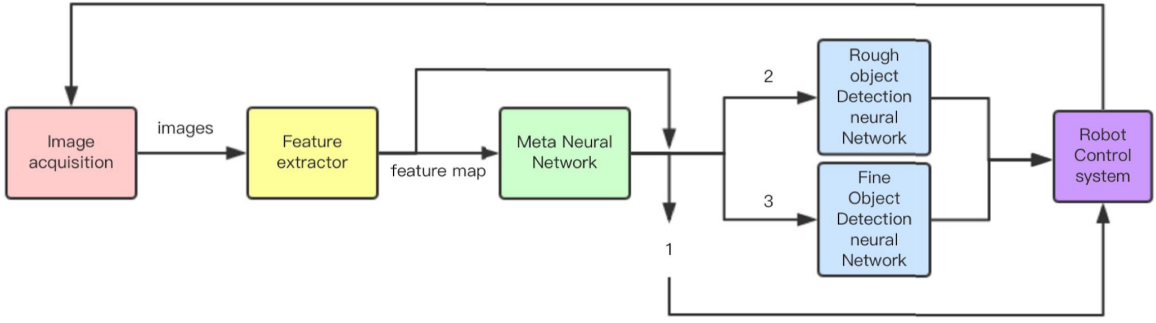

Fig. 2: Overall Structure of MAOD System

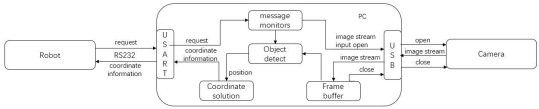

Fig. 3: The structure of image acquision subsystem

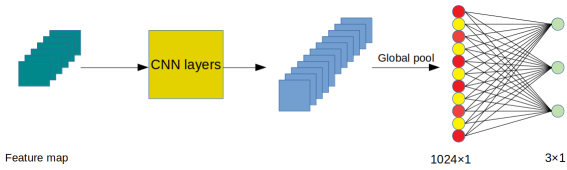

Fig. 4: Overall Structure of Meta Neural Network

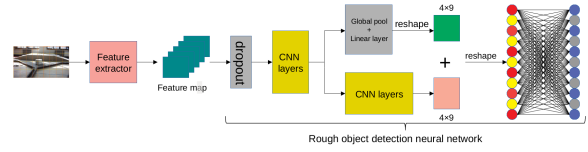

Fig. 5: Overall Structure of rough object detection neural network

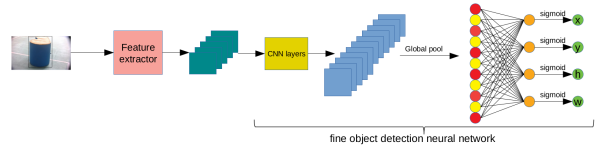

Fig. 6: Overall Structure of fine object detection neural network

### 3.5 Fine Object Detection Neural Network

This neural network here is to find the exact position of the object with a close view of this object. As shown in fig 6, the structure mainly contains the two components: CNN layers to extract the abstract feature based on the feature map generated by the feature extract and the linear layer, which predicts the position of the object. There is a sigmoid layer following the linear layer, whose goal is to condense the predicted value into range(0,1). The final output of the neural network is a vector which contains 4 elements, the center point of the object (x,y) and the width(w) and the height(h) of the object. The loss function of this network is:

$$loss = (x-x^{'})^2 + (y-y^{'})^2 + (w-w^{'})^2 + (h-h^{'})^2 \quad (3)$$

where $x, y, w, h$ is the predicted center point (x,y) of the object, w is the predicted value of width and h is the predicted height. $x^{'}, y^{'}, w^{'}, h^{'}$ are the corresponding target of the object.

### 3.6 Depthwise Separable Convolutions and Transfer learning

Obviously, the more trainable weights in a neural network, the more training data we need to train it. And the ability of a neural network is associated with its depth. The deeper neural network can extract more abstract features than a shallow neural network. Therefore, the question here is how to build a deeper network with small trainable weights. [23] points out that that the shallow layers of neural networks mainly extract simple common features of images, such as edge information, and the parameters of these layers do not differ significantly among different data sets after training. Therefore, here, we can make the shallow layers of the MAOD trained in Imagenet and fix the weights in this layer, so that we can deepen our network without extra trainable weights and the network still has the same effect. Specifically here, all the parameters in the feature extractor are frozen, as we can see from Table 1.This serves three purposes:

1) It can deepen the neural network without enlarging the trainable parameters, which can make the network easy to train.
2) It can help prevent overfitting. The lower layer of the neural network only extracts the lower feature of the image as we state in the introduction, therefore, with the fixed parameters can lower the neural network's ability to learning too much specific knowledge.
3) It makes the three neural network, meta neural network, rough neural network, and rough neural network, share the same feature extractor possible, which contribute a lot to improve the speed.

| module \ backbone | Mobilenetv2 | ShuffleNet | | |
|---|---|---|---|---|
| FE | M.conv2d<br>M.bottlenet1<br>M.bottlenet2<br>M.bottlenet3<br>M.bottlenet4 | SF.conv1<br>Maxpool<br>SF.stage2<br>SF.stage3 | | |
| MNN | M.bottlenet5<br>M.bottlenet6<br>M.bottlenet7<br>Global Pool<br>Linear Layer | SF.stage4<br>SF.stage5<br>Global Pool<br>Linear Layer | | |
| RODNN | Dropout<br>M.bottlenet5<br>M.bottlenet6<br>M.bottlenet7 | | Dropout<br>SF.stage4<br>SF.stage5 | |
| | Global Pool<br>Linear Layer<br>Reshape | conv1<br>conv2<br>conv3 | Global Pool<br>Linear Layer<br>Reshape | conv1<br>conv2<br>conv3 |
| | Plus<br>Linear Layer<br>Sigmoid | | Plus<br>Linear Layer<br>Sigmoid | |
| FODNN | M.bottlenet5<br>M.bottlenet6<br>M.bottlenet7<br>Global Pool<br>Linear Layer<br>Sigmoid | SF.stage4<br>SF.stage5<br>Global Pool<br>Linear Layer<br>Sigmoid | | |

Table 1: The table shows the specific structure of our neural network. Here the FE is short for feature extract, MNN meta neural network, RODNN rough object detection neural network and FODNN fine object detection neural network. The layers with M. means that these layers are defined in the mobilenetv2 neural network and they are initialized with the weights pretrained in Imagenet. And the layers with SF. means that these layers are defined in the ShuffleNet neural network and they are initialized with the weights pretrained in Imagenet.

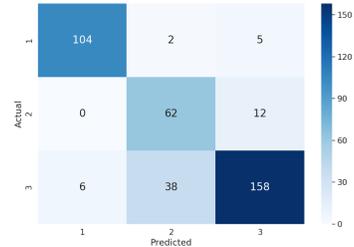

(a)

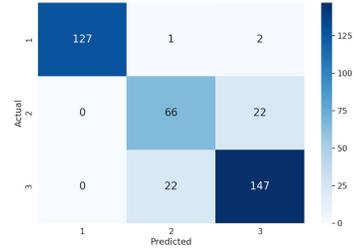

(b)

Fig. 7: (a) is the confusion matrix for meta neural network with mobilenetv2 as the backbone and (b) is the confusion matrix for meta neural network with shufflenet as the backbone. Because we randomly choose images from the dataset set as the test set in the experiment, the total number of each actual label is different in these two matrixes.

What's more, we only consider to use MobileNetv2[20] and ShuffleNet[24] as our backbone table 1. The main character of these neural networks is the use of depthwise separable convolution in these neural networks. This kind of convolution has some fascinating characters that we need to consider here.

First, compared with the traditional convolution method, this network can achieve nearly the same effect with much fewer parameters. Fewer parameters can easy the burden to train the neural network, just because we need to find the less optimal value for these parameters. This feature can help us to train our model with less dataset. Furthermore, depthwise separable convolution has a higher speed than traditional convolution. As a result, our system can process more frames in a limited time and this makes our robot more sensitive to the changes of the outside world.

## 4 Experimental results and discussion

To demonstrate the effectiveness of the proposed approach, we use the dataset which contains the different kinds of images perceived by the robot. This dataset contains three kinds of images corresponding to the situation illustrated in fig 1. The first situation, which contains no objects, has 607 images. Some of the images are the background of the scene and the others are the vague images, which obtained when the robot is moving its cameral to search for the object. Situation 2, which contains objects varied from 1 to 4 with a far view, has 452 images. There are different color objects here, including white, red and blue. Situation 3 has 328 images. All the objects and scenes in situation 3 are the same as situation 2, but with a close view. Meanwhile, the proposed approach is compared with other approaches object detection method. In addition, we will show our algorithm implement on the robot.

### 4.1 Experiments on Robot Vision Dataset

In this part, we use i5 8400 CPU in the computer to imitate the CPU on the robot to execute the experiment. Even

| Method | backbone | Meta | | st2 | | st3 | |
|---|---|---|---|---|---|---|---|
| | | accuracy | cpu time | F1 score | cpu time | F1 score | cpu time(s) |
| YOLOV3 | darknet | —— | —— | **79.0** | 0.259 s | 63.9 | 0.257 |
| Faster R-CNN | mobilenetv2 | —— | —— | 62.1 | 2.371 s | 63.6 | 2.292 |
| ours | mobilenetv2 | 87.8 | **0.039** | 73.2 | 0.122 s | 69.4 | 0.569 |
| **ours** | **ShuffleNet** | **88.9** | 0.042 | 74.5 | **0.038** | **98.0** | **0.114** |

Table 2: Compare of our method with YOLOV3 and Faster R-CNN. st2 is the situation where the robot is far from the object and st3 is the situation when the robot is closer to the object and it needs to refine its location to grasp the object

though the device is different from that in the robot, it can still tell the performance of our algorithm. Table 2 shows the result of our method with the shuffenet and mobilenetv2 as backbone compared with the YOLOV3 and Faster R-CNN. Here we use the F1 score and the average CPU time for each frame as our evaluation metrics:

$$Precision = \frac{T}{NP} \quad (4)$$

$$Recall = \frac{T}{NT} \quad (5)$$

$$F_1\ score = \frac{2 \times recall \times precision}{recall + precision} \quad (6)$$

$$CPU\ time = \frac{TT}{NF} \quad (7)$$

where $T$ is the number of predictions that are true, $NP$ denotes the number of all prediction, $NT$ denotes the number of targets, $TT$ is the total time needed for a sequence of input images and $NF$ denotes the number of images in this sequence. T In st2, we outperform the Faster R-CNN and YOLOV3 in both the F1 score and CPU time. Even though we are still worse in the F1 score compared with YOLOV3, our speed is still a big advantage. In YOLOV3, the speed is so slow that we cannot make the robot to react to the changes in the environment when we use this algorithm. And in the st3, the F1 score of our algorithm is nearly 1, which is much higher than YOLOV3 and Faster R-CNN. The accuracy here is important for the object to grasp the object. The reason for the low F1 score for YOLOV3 and Faster R-CNN is that these algorithms are striving to adapt to the two different situations. Our method successfully solve this problem using a meta neural network to choose different object detection. But the time for our method overall is not just the time of the object detection plus the meta neural network, because of all of them share the same feature extractor. The result illustrates that the embedding meta neural network only takes no more than 0.01s extra time for each image. Even with this extra time, we still have an obvious advantage over the other algorithms.

Fig 7 demonstrates the confusion matrix for the meta neural network. We can obtain from fig 7 that the meta neural network performs well to determine whether there are objects in the view of the robot. However, it struggles to distinguish situation 2 from situation 3, especially the robot is on the border of these two situations. Fortunately, it is no matter for the robot to choose fine object detection neural network or the rough object neural network when it is in the border in real practice. under this situation, the robot will move closer to the object, and then the meta neural network will have more and more confidence to determine the robot is in situation 3.

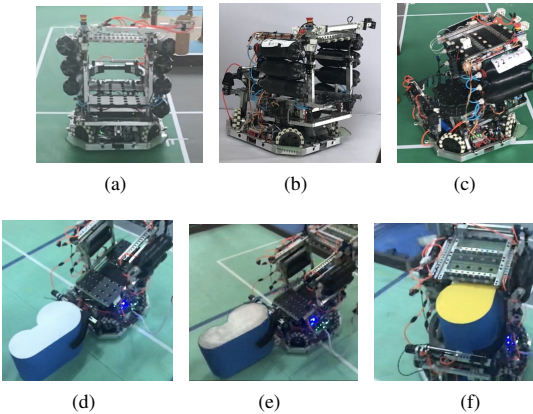

Fig. 8: (a),(b),(c) show our robot with different perspectives and (d),(e) are the situation where the robot is fetching the object. In (f), the robot is carrying the object to a destination.

### 4.2 Experiments on Real Robot

Fig 8 (a),(b),(c) give an overview of our robot. The robot consists of two main parts, the lower part is responsible for the movement of the robot and the upper part of the gripper. The frame of the chassis is a square in overall welded by a hollow square aluminum column of 20cm*40cm. Four Swedish wheels are placed at the four corners of the square, and the Swedish wheels are driven by motors. The robot relies on such a four-axis Swedish wheel system for motion, which can achieve linear or curved motion in any direction, thus ensuring the flexibility of the robot movement. The operation of the motors is controlled by a self-designed circuit board with an stm32 chip as the core. The trajectory is designed according to the planned motion route and the speed required for the motion, according to which, the codes are programmed and finally written to the chip. After powering the chip and the motor respectively, the chip can send different rotation commands to the motors according to the codes to control the rotation of the Swedish wheel system, so that the robot can move according to the expected route. In this process, the robot will constantly run the rough object detection algorithm to determine the fine route to get close to the object.

The gripper is located in the upper of the robot, which is made of carbon plate, controlled by the steering gear. The arm shaft can flip powered by the motors on the axis. After the robot approaches the appropriate distance to the object placed on the ground, the gripper flips to the outside to grasp it, and then flips it to the inside to feed the object into the robot. In fig 8 (d)(c), the robot is trying to grasp the robot. Before this behavior, fine object detection is used to find a fine location for this robot, so that the robot can precisely catch the object.

In actual control, the chassis circuit board acts as the master and supervises the overall action flow of the robot. The circuit board that controls the upper structure acts as a co-master and works according to the instructions of the master. The master controls the robot to move as the planned route acquiring the position information of the current robot in real-time. The command will be sent to the co-master through the CAN network by the master if at the expected position, and the co-master controls the movement of the gripper. Thus, such a system is able to allow the robot to perform as expected at different locations on the planning path.

### 4.3 Conclusion

The proposed system is mainly used for the robot in the task that the robot needs to fetch an object in a far distance. Out system can successfully address the problem that the view of the robot changes much when the robot gets closer to the object gradually. What's more, our method has a high speed which makes the robot sensitive to the changes in our environment.